\documentclass[preprint,groupedaddress,11pt]{revtex4-1}

\usepackage{amsmath}    
\usepackage{graphicx}   
\usepackage{verbatim}   
\usepackage{color}      
\usepackage{subfigure}  
\usepackage{hyperref}   
\usepackage{url}
\usepackage{amssymb}
\newcommand{\movies}{\emph{MOVIES }}
\DeclareMathOperator*{\argmax}{arg\,max}
\usepackage{bm}
\newcommand{\vect}[1]{\boldsymbol{\mathbf{#1}}}

\usepackage{multirow}

\begin{document}

\title{Machine Learning Sentiment Prediction based on Hybrid Document Representation}
\author{Panagiotis Stalidis, Maria Giatsoglou, Konstantinos Diamantaras\footnote{kdiamant@it.teithe.gr}}
\affiliation{Department of Information Technology, TEI of Thessaloniki, GR-57400 Thessaloniki, Greece}
\author{George Sarigiannidis, Konstantinos Ch. Chatzisavvas\footnote{kchatz@msensis.com}}
\affiliation{mSensis S.A., VEPE Technopolis, Bld C2, GR-57001 Thessaloniki, Greece}


\begin{abstract}
Automated sentiment analysis and opinion mining is a complex process 
concerning the extraction of useful subjective information from text. 
The explosion of user generated content on the Web, especially the fact 
that millions of users, on a daily basis, express their opinions on products and services 
to blogs, wikis, social networks, message boards, etc., render 
the reliable, automated export of sentiments and opinions from unstructured text 
crucial for several commercial applications. 
In this paper, we present a novel hybrid vectorization approach for textual resources that combines  
a weighted variant of the popular Word2Vec representation (based on Term Frequency-Inverse Document Frequency) 
representation and with a Bag-of-Words representation 
and a vector of lexicon-based sentiment values. 
The proposed text representation approach is assessed through the application of several machine learning classification algorithms  
on a dataset that is used extensively in literature for sentiment detection. 
The classification accuracy derived  through the proposed  hybrid vectorization approach is higher than
when its individual components  are used for text represenation, and comparable with state-of-the-art 
sentiment detection methodologies. 
\end{abstract}

\maketitle

\section{Introduction}
\label{Introduction}

\emph{Sentiment Analysis} (often referred to as \emph{Opinion Mining}) 
involves the application of Natural Language Processing, text analytics, 
and computational linguistics methods and tools in order to extract 
subjective information from text. It is applied on a variety of 
textual sources that typically carry peoples' opinions, emotions, 
and evaluations on specific \textit{entities}, such as individuals, 
events or topics (for instance, reviews of movies, books, products, etc.). 
Sentiment Analysis aims mainly at \emph{identifying the sentiment content} of the  
textual resource under inspection, and subsequently \emph{estimating its polarity} (positive/negative). 
Related tasks are \emph{question answering} (recognizing opinion oriented questions) 
and \emph{summarization} (accounting for multiple viewpoints). The sentiment analysis process 
can be carried out on different granularity levels: at the level of the \textit{document}, \textit{sentence} or \textit{aspect}. 
\emph{Document-level} Sentiment Analysis aims at classifying a multi-sentence document 
in terms of the polarity of the opinion expressed within it. 
This approach assumes that the whole document expresses opinions on a 
single entity (e.g. a single topic), and is not applicable to documents that refer 
to multiple entities~\cite{pang2002thumbs}. \emph{Sentence-level} or 
\emph{phrase-level} Sentiment Analysis seeks to classify the sentiment expressed 
in a single sentence, and characterize the sentence as positive, negative, or neutral. 
Liu \citeyear{liu2012sentiment} argues that there is no fundamental difference 
between document- and sentence-level classification, since sentences 
are simply short documents. \emph{Aspect-level} Sentiment Analysis aims to extract 
sentiments expressed with respect to certain aspects of the entities. 
This is necessary in many applications, especially when products or services are evaluated. 
For example, the sentence ``\emph{The voice quality of this phone is not good, 
but the battery life is long}", expresses a negative sentiment towards the quality 
of the product but a positive sentiment regarding its durability.
Typically, Sentiment Analysis is applied on text corpora containing product reviews. 
However, it can also be applied on news articles~\cite{xu2012identifying} or on documents 
which are generated in social networks and microblogging sites. 
In the latter case the aim is to extract the public opinion on various topics ranging from
stock markets~\cite{hagenau2013automated} to political debates 
\cite{maks2012lexicon}.
 

The two main approaches for automated sentiment extraction are the 
\emph{Lexicon-based} approach and the \emph{Machine Learning} approach.  

Lexicon-based Sentiment Analysis involves the estimation of the sentiment of a textual resource 
from the semantic orientation of the words or phrases it includes~\cite{turney2002thumbs}. 
It has primarily focused on using adjectives as indicators of the semantic orientation 
of text~\cite{hatzivassiloglou1997predicting,wiebe2000learning,hu2004mining,taboada2006methods} 
or adjectives and adverbs combined~\cite{benamara2007sentiment}.
Lists of such terms together with their sentiment orientation values are collected 
into dictionaries, and then the sentiment of a textual resource is estimated through the 
aggregation of the values of the lexicon terms found in the document. 
Dictionaries for lexicon-based approaches can be created manually~\cite{taboada2011lexicon,tong2001operational} 
or automatically~\cite{hatzivassiloglou1997predicting,turney2002thumbs,turney2003measuring} 
using seed words to expand the list of words. 
Since one of the steps in lexicon-based approaches
is to annotate text with part-of-speech tags, the aspects that the sentiment is targeted upon 
can be easily detected.

Machine Learning (ML) Sentiment Analysis involves supervised training of classifier models 
using labeled instances of documents or sentences~\cite{pang2002thumbs}. 
A number of features are extracted from the text, such as words (unigrams), 
groups of successive words (n-grams), or even numeric features, such as the 
number of nouns, verbs, or adjectives, included in the text. 
Then, these features are represented by vectors which are fed into suitable ML 
algorithms for deriving the classification model. Many text classifiers have been proposed in
literature, including Decision Trees, Na{\"i}ve Bayes, Rule Induction, Neural Networks, Clustering K-Nearest Neighbours, 
and Support Vector Machines (SVM). One of the earliest works following the ML approach has been 
proposed by Pang et al.~\citeyear{pang2002thumbs}. The authors employed a bag-of-features representation approach 
on movie reviews using unigrams, bigrams, adjectives and the position of words as features. 
In order to predict sentiment they used a number of standard ML classifiers concluding that the best 
performance is achieved when the unigrams are used in combination with an SVM classifier.
Whitelaw et al.~\citeyear{whitelaw2005using} augmented the bag-of-words approach with shallow 
parsing that extracts opinion phrases, classified into attitude types derived from appraisal theory~\cite{martinlanguage}. 
An SVM was applied on the feature vectors comprising word frequencies and the percentage 
of appraisal groups with particular orientations. They achieved 90.2\% accuracy classifying 
the movie reviews corpus introduced by Pang et al.~\citeyear{pang2002thumbs}.  
Ye et al.~\citeyear{ye2009sentiment} used sentiment classification techniques for classifying 
online reviews for travel destinations. 
They used the frequency of words instead of word presence to represent a document, and the
Information Gain measure to select the feature set. They compared the performance of SVM, 
Na{\"i}ve Bayes and a character-based N-gram model on the classification of travel destination reviews. 
SVM was found to outperform the other two classifiers with an accuracy peak of $\sim$86\% 
when the training corpora contained 700 reviews. Prabowo and Thelwall~\citeyear{prabowo2009sentiment} 
used a hybrid classification scheme employing a cascade of rule-based classifiers in combination 
with an SVM-based approach to separate the two classes (positive/negative). The hybrid 
classifier was tested on movie reviews, product reviews and social media comments (from MySpace) 
yielding an F-measure score between 72.77\% and 90\%.

This paper introduces a hybrid method that combines both lexicon-based features and machine learning 
for sentiment classification.
We propose a novel document representation approach that combines the Bag-of-Words representation 
with a TF-IDF weighted Word2Vec representation together with a vector of lexicon-based sentiment values. 
Several supervised machine learning methods are then trained using the derived
representation vectors as inputs and the polarity values as targets, and are comparatively evaluated based 
on their performance. 
For our simulation experiments we selected a well-known movie reviews corpus, so that our results can be 
easily compared against other methods.

The rest of the paper is organized as follows.
In Section~\ref{sec:vector} we describe the sentence vector representation methodology, and
in Section~\ref{sec:classmodels} we outline the machine learning models used for training the system.
Then, in Section~\ref{sec:results} we evaluate the performance of our proposed approach 
based on a series of experiments, and compare it with competitive methods. 
Finally, in Section~\ref{sec:discussion} we discuss the benefits and limitations of the proposed approach.


\section{Vector Representation of Sentences}
\label{sec:vector}


Efficient vector representations of 
documents is essential to any algorithm that performs Sentiment Analysis.
Typically, such representations have one of the two following properties:
\begin{itemize}
	\item[(a)] ability to capture semantic similarities between documents, 
	i.e. paragraphs of similar meaning are coded by vectors which are close to each other in  Euclidean space; 
	\item[(b)] ability to capture sentiment polarity or emotional content. 
\end{itemize}
Unfortunately, representations that have the first property fail to capture sentiment.
On the other hand, lexicon-based approaches that have the second property fail to capture semantic similarity 
between documents.
The motivation behind our proposed approach is our belief that a combination of both properties is beneficial, 
if not necessary, for sentiment analysis.

The first property facilitates the improvement of the generalization capability for any ML model that uses 
these representations to classify documents in terms of their sentiment polarity or emotional content.
The Bag-of-Words (BoW) approach~\cite{salton1986introduction}, for instance, represents a document using 
the frequency of occurrence of each word, disregarding the word order, and often considering only terms from a given dictionary or corpus.
Semantic similarity between documents can be captured by comparing their term frequency profiles (i.e. representation vectors).
However, BoW fails to capture similarity,in the case of synonyms.
For example, the sentences: ``\emph{He came late to school}'' and ``\emph{He delayed coming to class}'' 
have quite different BoW representations although they have very similar meanings.
The Word2Vec model \cite{mikolov2013efficient}, described in more detail below, 
tries to overcome this limitation by representing words with vectors such that words appearing frequently 
in similar contexts have close representations in the Euclidean space.
Even so, such semantics-based representations alone, ignore the sentimental 
value of the terms included in the document.

The second property can be achieved, for example, by the use of a lexicon where terms are assigned 
scores corresponding to their sentiment orientation. Early approaches that are characterized by  
only this property are called \emph{lexicon-based} methods~\cite{taboada2011lexicon}.
They suffer from low coverage, i.e. the fact that many sentences may not contain any terms from the 
lexicon and, thus, their sentimental orientation cannot be evaluated. To mitigate the disadvantages 
of both approaches, we propose to use lexicon-based representations in complementary fashion 
to the semantics-based representations. Thus, the combined representation will  have both properties 
described above. 

Motivated by the above discussion, we propose the \textbf{Hybrid Weighted Word2Vec} (HWW2V) 
text representation approach, which is a concatenation of:
\begin{enumerate}
	\item[(a)] \emph{Bag-of-Words} representation,
	\item[(b)] \emph{Weighted Word2Vec} representation, and
	\item[(c)] \emph{Sentiment lexicon-based} representation. 
\end{enumerate}
The next subsections provide the details for each individual representation approach.

\subsection{Morphological Processing}

Prior to extracting the features that are fed into the classifier, 
a number of preprocessing steps are taken as described below:

\begin{itemize}
	\item \emph{Tokenization}: Split the text into smaller parts mainly words.
	\item \emph{Contractions}: Replace contractions with two tokens, e.g., 
	\textsf{can't} is replaced by \textsf{can} and \textsf{not}.
	\item \emph{Negations}: Negations are treated by adding artificial words \cite{wiegand2010survey}. 
	For example, if a word $x$ is preceded by a negation word, then a new feature \textsf{negated\_x} is created. 
	Since, as suggested by \cite{pang2002thumbs}
	the scope of negation cannot be properly modelled, every word is replaced until the end of the sentence.
	\item \emph{Stopword Removal}: Remove common words which convey no sentiment or semantics, 
	such as ``a'', ``the'', ``to'', etc.
\end{itemize}

\subsection{Bag-of-Words Representation}

The dictionary used in the BoW representation is the set of all words in the document corpus after removing 
the stop words. Furthermore, all words were turned to lower-case but no lemmatization or stemming was applied. 
We used the Term Frequency-Inverse Document Frequency (TF-IDF) representation of each document 
where the term frequency was forced to be binary (i.e. a given word is present or not) 
and only the document frequency varied.

\subsection{Weighted Word2Vec Representation}

Mikolov et al.~\citeyear{mikolov2013efficient} proposed the \emph{Word2Vec} model that embeds semantic 
information in a vector space representation of words. Its key idea is the capturing  
of the context of words by using ML approaches such as Neural Networks. Word2Vec predicts the 
occurrence of a word given a window of previous and subsequent words. It incorporates two different architectures for the 
computation of vector representations of words from large datasets:
\begin{itemize}
	\item[(a)] Continuous Bag-of-Words model (CBOW): predicts a word when the surrounding 
	words are given. It is much faster than the Skip-gram model and slightly more accurate for 
	frequent words;
	\item[(b)] Skip-gram model (SG): predicts a window of words when a single word is known. 
	It operates well with a small amount of training data representing accurately even rare words
	and phrases.
\end{itemize} 
The learned vectors explicitly encode many linguistic patterns, for example, vectors encoding words 
with similar meanings are close in the Euclidean space. Also, many of these patterns can be represented 
as linear translations~\cite{mikolov2013linguistic}. Following these successful techniques, research 
interest has been extended to the development of models that go beyond word-level to achieve sentence-level 
representations~\cite{zanzotto2010estimating,mitchell2010composition,grefenstette2013multi,yessenalina2011compositional,socher2013recursive,le2014distributed}. 
Probably, the most straightforward approach to derive document-level representations is to simply take 
the average of the vectors of all words contained in the document, according to the Word2Vec embedding (i.e. \emph{mean Word2Vec}).
Doc2Vec is a more sophisticated approach, which modifies the original Word2Vec algorithm to the
unsupervised learning of continuous representations for larger blocks of text, such as sentences, 
phrases, paragraphs or entire documents~\cite{le2014distributed}.

\paragraph{Proposed approach:} The \emph{Weighted Word2Vec}, introduced here, is a modification of \emph{mean Word2Vec}, 
where we represent a document by the \emph{weighted average} of the representations of all its words 
as they are obtained by the Word2Vec model. For each word $i$ in the dictionary we compute its vector representation 
$\vect{v}_i$ by using the \emph{CBOW} model and we construct a \emph{dictionary matrix} 
\begin{align*}
	\vect{V}\in\mathbb{R}^{N\times B},
\end{align*} 
where $N$ is the dimensionality of each word representation and $B$ is the size of the dictionary. 
Each term \textit{weight} should reflect its "importance" within the document, and thus it is directly proportional to its 
\emph{Term-Frequency} normalized by the \emph{Inverse-Document-Frequency}, i.e. the TF-IDF measure.
For each document $j$ and word $i$ we calculate the TF-IDF, as
\begin{align}
	w_{ij}=f_{ij}\,  \log \dfrac{D}{D_i}
\end{align}
where $f_{ij}$ is the frequency of the $i$-th word in document $j$, $D$ is the total number of documents in the corpus, 
and $D_i$ is the number of documents containing the $i$-th word. Then the vector 
representation $s_j$ of the $j$-th document is the sum of the words that appear in it weighted by their TF-IDF values:
\begin{align}
	\vect{s}_j=\sum_{i} w_{ij}\, \vect{v}_i
\end{align}

\subsection{Sentiment Lexicon-based Representation}
The Word2Vec-based representations described in the previous paragraphs do not contain sentiment information 
about the words included in the document. We wish to study the advantages of augmenting this representation 
by including the semantic orientation of words based on a lexicon that associates a set of terms with their sentimental 
polarity expressed as a numerical value. This approach, of course, assumes that semantic orientation is independent 
of context. The lexicon can be created either manually, such as the General Inquirer lexicon~\cite{stone1966general}, 
or semi-automatically, utiling resources like WordNet~\cite{hu2004mining,esuli2006sentiwordnet}.
Our implemented model makes use of the semi-automatically created sentiment lexicon 
SentiWordNet (SentiWordNet) which has been applied in different opinion related tasks, 
i.e., for subjectivity analysis and sentiment analysis, with promising results. 

SentiWordNet extends the WordNet lexicon of synonyms adding sentiment values for each synset 
(i.e., sets of cognitive synonyms) in the collection by means of a combination of linguistic and statistic classifiers. 
SentiWordNet assigns to each synset of WordNet three sentiment scores, based on its \textit{positivity}, 
\textit{negativity} and \textit{objectivity}, respectively, with the sum of these scores being always 1. 
The model we implemented using SentiWordNet, calculates a positive, objective and negative score 
for each token encountered in the corpus, by averaging the values of all synsets in SentiWordNet 
that have the same text representation. Tokens that are used after a negation, i.e. \textsf{NOT\_good}, 
have their polarity reversed. In order to anticipate words that are not included in the dictionary, 
a fourth feature was added, denoting unknown sentiment. The sentiment scores for all tokens 
(including their reversed forms) are represented in the \textit{sentiment matrix}.

Since a token generally does not pass its sentiment in another sentence (apart from the one it resides in), 
we first split each review in sentences. Then, for each sentence we form a vector as in a BoW model, 
and we aggregate its overall sentiment representation by calculating the dot product of the sentiment matrix 
with the sentence vector. 
To aggregate the sentiment at the document level, we calculate the mean sentiment representation
across all the document's sentences. This process extracts four features for each document:
\begin{enumerate}
	\item a positive sentiment value, 
	\item an objective value, 
	\item a negative sentiment value, and
	\item a value indicating the percentage of words that are not included in SentiWordNet. 
\end{enumerate}

\noindent \emph{Example}: 


\begin{center}
\begin{tabular}{l |c| cccc}
	\hline
	 Text & Sentence Vector & Positive & Objective & Negative & Unknown
\\
	\hline \hline
	- 		&1 & 0 & 0 & 0 & 1 \\
	. 		&1 & 0 & 0 & 0 & 1 \\
	bad 	& 1 & 0 & 0.2 & 0.8 & 0 \\
	between & 1 & 0 & 1 & 0 & 0 \\
	dead    & 1 & 0.1 & 0.3 & 0.6 & 0 \\
	man     & 1 & 0 & 1 & 0 & 0 \\
	room    & 1 & 0 & 1 & 0 & 0 \\
	smell    & 1 & 0.2 & 0.6 & 0.2 & 0 \\
	smt       & 1 & 0 & 0 & 0 & 1 \\
	so         & 2 & 0 & 1 & 0 & 0 \\
	towels   & 1 & 0 & 1 & 0 & 0 \\
	wardrobe & 1 & 0.1 & 0.9 & 0 & 0 \\
	wet       & 1 & 0.1 & 0.8 & 0.1 & 0 \\
	\hline
\end{tabular}
\end{center}

\noindent The inner product of the sentence vector and the sentiment matrix 
\begin{equation*}
	\begin{pmatrix}
	1 & 1 & 1 & 1 & 1 & 1 & 1 & 1 & 1 & 2 & 1 & 1 & 1  \\
	\end{pmatrix}
	\cdot
	\begin{pmatrix}
	0 & 0 & 0 & 1 \\
	0 & 0 & 0 & 1 \\
	0 & 0.2 & 0.8 & 0 \\
	0 & 1 & 0 & 0 \\
	0.1 & 0.3 & 0.6 & 0 \\
	0 & 1 & 0 & 0 \\
	0 & 1 & 0 & 0 \\
	0.2 & 0.6 & 0.2 & 0 \\
	0 & 0 & 0 & 1 \\
	0 & 1 & 0 & 0 \\
	0 & 1 & 0 & 0 \\
	0.1 & 0.9 & 0 & 0 \\
	0.1 & 0.8 & 0.1 & 0
	\end{pmatrix}
	=
	\begin{pmatrix}
	0.036 & 0.671 & 0.121 & 0.171
	\end{pmatrix}
\end{equation*}

\noindent Thus, the vector representation of the text is 
\begin{center}
\begin{tabular}{cccc}
	\hline
	Positive & Objective & Negative & Unknown  
	\\
	\hline \hline
	0.036 & 0.671 & 0.121 & 0.171 \\
	\hline
\end{tabular}
\end{center}


\section{Machine Learning Classification Models}
\label{sec:classmodels}

The document representations resulting from the concatenation of the BoW, Weighted Word2Vec, 
and Sentiment Lexicon-based vectors are used to train a supervised classifier to learn the documents' sentiment polarity.
In this work we have evaluated a number of popular machine learning methods as detailed below.

\subsection{Na{\"i}ve Bayes}
The probability of a document $d$ being in class $c$ is computed as
\begin{align}
 	P(c | d) \propto P(c) \prod_{1\leq k \leq n_d} P(t_k | c) 
\end{align}
where $(t_1,t_2,\ldots,t_{n_d})$ are the tokens in the vocabulary that appear in 
$d$ and $n_d$ is the number of the tokens. 
Classification is performed by choosing the maximum a posteriori (MAP) estimate:
\begin{equation}
	C_{\rm map}=\argmax_{c\in C} P(c | d) 
\end{equation}
In order to avoid floating point underflow we use the log probability 
\begin{align*}
	\log P(c | d) = \log P(c) + \sum_{1\leq k\leq n_d} \log P(t_k | c)
\end{align*}	
Ensemble estimates are used for the values $P(c)$ and $P(t_k | c)$. 
One complication is that if a term $t_k$ never appears in class $c$, the value $P(t_k | c)$ is zero and 
due to the multiplicative rule in Na{\"i}ve Bayes, the overall probability will be zero, independent of how strong evidence 
other terms may offer. We avoid this problem by adding $1$ to all term counts, thus modifying the equation 
to estimate the conditional probability to
\begin{align}
	P(t_k | c)=  \frac{N_{c t}+1}{\sum_{t' \in V} (N_{c t'}+1)}  = \frac{N_{c t}+1}{(\sum_{t'\in V} N_{c t'}) +B} 
\end{align}
where $B$ is the number of terms in the vocabulary $V$.

\subsection{Maximum Entropy}

Considering any feature $x_i (d,c)$ of a document $d$ in class $c$, the maximum entropy principle restricts the model distribution 
to have the expected value of this feature equal to the average obtained from the training data, $D$
\begin{align}
	\frac{1}{|D|}\, \sum_{d\in D} x_{i}(d,c)=\sum_{d} P(d) \sum_{c} P(c|d)\,x_{i}(d,c)
	\approx \frac{1}{|D|}\sum_{d\in D}\sum_{c} P(c|d)\,x_{i}(d,c) 	
\end{align}
When constraints are imposed in this fashion, it is guaranteed that a unique distribution exists 
that has maximum entropy~\cite{nigam1999using}. 
Moreover, it has been shown~\cite{pietra1997inducing}
that the distribution is always of the exponential form
\begin{align}
	P(c|d)=\frac{1}{Z(d)}\,{\rm exp}\left( \sum_{i}\lambda_{i}\,x_{i}(d,c)\right)
\end{align}
where $\lambda_{i}$ is a parameter to be estimated and $Z(d)$ is the \emph{partition function} 
that ensures that $P(c|d)$ is a proper probability.
When the constraints are estimated from labeled training data, the solution to the ME problem 
is also the solution to a dual maximum likelihood problem for models of the same exponential form. 
Additionally, it is guaranteed that the likelihood surface is convex, having a single global maximum and no local maxima. 
A possible approach for finding the maximum entropy solution would be to guess any initial exponential distribution 
of the correct form as a starting point and then perform hill climbing in likelihood space. 
Since there are no local maxima, this will converge to the maximum likelihood solution for exponential models, 
which will also be the global maximum entropy solution.

\subsection{Support Vector Machines}

Support Vector Machines (SVM)~\cite{cortes1995support} are called maximum margin classifiers, since, 
in the case of a separable problem, the obtained separating surface maximizes 
the distance of the closest pattern from it. 
This is achieved by solving the following optimization problem:
Maximize
\begin{align}
	L(\vect{a}) = \sum_{i} a_{i} - \dfrac{1}{2} \, 
	\sum_{i,j} a_{i} a_{j} y_{i} y_{j} \, K(\vect{x}_{i},\vect{x}_{j})
\end{align}
under the conditions
\begin{align*}
	0\leq a_{i}\leq C \quad \text{and}\quad \sum_{i} a_{i}y_{i}=0
\end{align*}
where $y_{i}\in\{-1,1\}$ is the class label for document $i$, $\vect{x}_i$ is the vector representation of document $i$, 
$C$ is a user parameter specifying how important 
the misclassification penalty is, and $K(\cdot,\cdot)$ is a kernel function
(polynomial, Radial Basis Function (RBF), hyperbolic tangent, etc.). 
After solving the above optimization problem the optimal classifier is 
\begin{align}
	f(\vect{x})=\sum_{i} a_{i}y_{i}\,K(\vect{x}_{i},\vect{x})
\end{align}


\section{Experiments}
\label{sec:results}


\subsection{Experimental Methodology}

In all experiments reported below we evaluated the classification models using a 10-fold cross-validation protocol 
leaving 10\% of the patterns out for testing.

\begin{figure}[h!]
	\centering
	\includegraphics[scale=0.5]{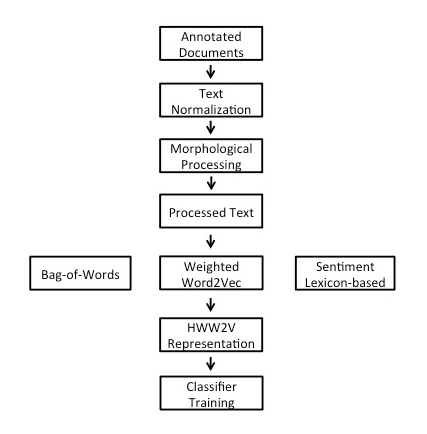}
	\caption{Experimental set-up}
	\label{fig:experimental_methodology}
\end{figure}

Figure \ref{fig:experimental_methodology} describes the steps used in our experiments. 
First, the documents are pre-processed in order to handle negations, contractions, stop-words, etc., 
as described in the next subsection.
Then the processed documents are represented in three different ways using the BoW, Weighted Word2Vec, 
and Sentiment lexicon-based vectors as described in Section \ref{sec:vector}. 
Additionally, all three vectors are combined together in the ``HWW2V'' representation. 

All representations, the three individual and the hybrid one, were tested separately using different types of shallow learning classifiers: 
Na{\"i}ve Bayes, Maximum Entropy, SVM with linear kernel, and SVM with RBF kernel (see Section \ref{sec:classmodels}) based on the accuracy 
of the classification model on the sentiment detection task. The purpose of this experiment is to investigate the potential improvement 
offered by the hybrid representation.


\subsubsection{The Movie Reviews Dataset} 

To test our method we use the popular RT movie reviews data set\footnote{\url{http://www.cs.cornell.edu/people/pabo/movie-review-data/rt-polaritydata.tar.gz}}
introduced by Pang and Lee~\citeyear{pang2005a}.
It is a collection of 10662 processed sentences/snippets from movie reviews written in English 
by users in the Rotten Tomatoes website. 
The collection contains 5331 positive and 5331 negative reviews.

\subsubsection{Machine Learning Algorithms}

As mentioned above, we employed both shallow and deep learning models for the classification stage.
In the case of shallow models the following popular algorithms were tested:
(a) Na{\"i}ve Bayes, (b) Maximum Entropy, (c) SVM with linear kernel, and (d) SVM with RBF kernel.
For all of the above methods we used Python's \emph{scikit-learn} machine learning implementation\footnote{\url{http://scikit-learn.org}}. 
For Na{\"i}ve Bayes, we used the MultinomialNB class which implements the Na{\"i}ve Bayes 
algorithm for multinomially distributed data, and is one of the two classic Na{\"i}ve Bayes variants 
used in text classification. For Maximum Entropy, we used the Logistic Regression class with the liblinear solver. 
In both cases, we examined if scaling the input vectors to unit form is beneficial for the prediction process. 
We also examined how the removal of terms with high and low document frequencies affects the accuracy of the classifiers.
For SVM, we used the SVC class implementation with both a linear kernel and an RBF kernel. 
In the case of the RBF kernel, two parameters must be considered, $C$ and $\gamma$. 
Parameter $C$, common to all SVM kernels, trades-off misclassification of training examples 
against simplicity of the decision surface. A low $C$ makes the decision surface smooth, 
while a high $C$ aims at classifying all training examples correctly. Parameter $\gamma$ defines how much influence 
a single training example has. The larger $\gamma$ is, the closer other examples must be to be affected.

The performance of the models is evaluated based on the average test-set accuracy after 10-fold cross-validation experiments.
\begin{eqnarray*}
	\text{accuracy}      &=& \dfrac{\text{TP+TN}}{\text{TP+TN+FP+FN}} 
\end{eqnarray*}
where TP and FP are the numbers of true and false positives, and TN and FN are the numbers of 
true and false negatives, respectively.


\subsection{Experimental Results}

The results of our experiments using the shallow learning classifiers are summarized in Table~\ref{tab:results_shallow_learning}.

\begin{table}
	\centering
		\caption{Classification performance using different document representations: 
	(a) only Sentiment Lexicon-based vectors, 
	(b) only Weighted Word2Vec vectors with dimension=100 (top row) and dimension=300 (bottom row), 
	(c) only BoW vectors, and  
	(d) HWW2V representation with Weighted Word2Vec vectors of dimension=100 (top row) and dimension=300 (bottom row). 
	All numbers in parentheses indicate execution times in seconds for one fold of the cross-validation experiment.}
	\begin{tabular}{p{3.3cm}p{2.5cm}p{2.5cm}p{2.3cm}p{2.8cm}}

	\hline 
	Method	& Sentiment Lexicon-based
	& Weighted Word2Vec & 
	BoW 
	& HWW2V	\\
	\hline \hline
	\multirow{2}{*}{Na{\"i}ve Bayes}
		& \multirow{2}{*}{0.570 (0.003)}
		& 0.541 (0.013)
		& \multirow{2}{*}{0.781 (0.012)}
		& 0.781 (0.034)
		\\
		&
		& 0.727 (0.030)
		& 
		& \textbf{0.785} (0.067) \\
	\hline
	\multirow{2}{*}{Maximum Entropy}
		& \multirow{2}{*}{0.582 (0.009)}
		& 0.623 (0.707)
		& \multirow{2}{*}{0.773 (0.095)}
		& 0.784 (0.842)
		\\
		&
		& 0.772 (40.6)
		&
		& \textbf{0.791} (125.7)\\
	\hline
	\multirow{2}{*}{SVM (linear)}
		& \multirow{2}{*}{0.582 (0.254)}
		& 0.620 (2.604)
		& \multirow{2}{*}{0.772 (0.299)}
		& 0.783 (2.590)
		\\
		&
		& 0.772 (349.8)
		&
		& \textbf{0.792} (376.1) \\
	\hline
	\multirow{2}{*}{SVM (RBF)}
		& \multirow{2}{*}{0.582 (5.02)}
		& 0.629 (22.54)
		& \multirow{2}{*}{0.776 (28.21)}
		& 0.783 (62.69)
		\\
		&
		& 0.777 (136.7)
		&
		& \textbf{0.796} (3,006.5) \\
	\hline
	\end{tabular}
	\label{tab:results_shallow_learning}
\end{table}

The results achieved using only the vectors obtained by the sentiment lexicon (Sentiment lexicon-based representation) 
are presented in the second column of Table~\ref{tab:results_shallow_learning}.
The obtained accuracies are low, regardless of the classifier, implying that the features are not very 
successful in capturing the document sentiment polarity.
This can be explained by the fact that SentiWordNet is a general sentiment dictionary 
which captures an overall (broad) sentiment for each word.

The third column of Table~\ref{tab:results_shallow_learning} presents the results obtained using only the 
Weighted Word2Vec representation vectors as features.
We note that the SVM model with RBF kernel slightly outperforms the other classifiers obtaining accuracy around 78\% 
for vectors with dimension 300.
The dimensionality of the vectors is an issue since the increase of the dimensions from 100 to 300 results to a substantial 
improve of the accuracy (14.8\%-18.6\%).
Almost the same results are obtained using the BoW approach for representing the sentences.
In this case, the Na{\"i}ve Bayes classifier yields the best performance with 78\% accuracy while the SVM/RBF model is a close second.
Both BoW and Weighted Word2Vec representations seem to offer the same discriminative power regarding sentiment polarity.

The HWW2V representation 
yields the results displayed in the last column of Table~\ref{tab:results_shallow_learning}.
It is clearly seen that HWW2V assists 
all four classifiers achieve better performance compared to using each representation separately.
The best performance is achieved for the SVM model with RBF kernel, obtaining accuracy slightly lower than 80\%.
For each classifier the performance is improved between 0.5\% (in the case of Na{\"i}ve Bayes) and 2.5\% (in the case of SVM with linear kernel) 
compared to the second best approach (BoW representation).

Our initial hypothesis was that lexicon-based features can not accurately capture refined characteristics and contextual cues 
that are inherent in the human language, mainly because people often express 
their emotions and opinions in subtle ways. Thus, we proposed a hybrid representation approach that combines 
sentiment lexicon-based features with word embedding-based approaches (such as Weighted Word2Vec), which 
try to capture semantic and syntactic features of words out of document collections 
in a language independent process. Our experiments showed that hybrid approaches can take advantage of both 
lexicon-based features and word embedding learning approaches to derive more accurate sentiment prediction results.



\section{Discussion}
\label{sec:discussion}

We propose a hybrid methodology that combines TF-IDF weighted Word2Vec representation 
and BoW representation with sentiment lexicon-based values for sentiment analysis in unstructured text. 
The word embedding representation, based on Word2Vec, provides semantic and syntactic coding, 
refined further in Weighted Word2Vec with TF-IDF weighting that augments the value of frequent words. 
The Sentiment Lexicon-based representation provides sentiment and emotion information for the document under inspection. 
The hybrid approach monetizes the merits of both representations, and it is 
customizable for any language as long as a sentiment lexicon is available.   

Our methodology is assessed through the application of several classifiers on the \movies dataset that has been used extensively 
in literature for the evaluation of the various methodologies proposed for sentiment analysis. 
In Table~\ref{review} we summarize the results of the application of different methods along with the performance 
of our approach, HWW2W.

\begin{table}[h]
	\centering
	\caption{Comparison of different methods on the \movies dataset. All results in this table, 
	except for the last row, are taken from \protect\cite{kim2014convolutional}.}
	\begin{tabular}{lcp{7cm}}
	\hline
	
	Method		& Test Accuracy		& Reference
	\\ \hline \hline
	CNN-rand			& 0.761		& \multirow{4}{*}{\cite{kim2014convolutional}} \\
	CNN-static			& 0.810		&  \\
	CNN-non-static		& 0.815		&  \\
	CNN-multichannel	& 0.811		&  \\ 
	\hline
	RAE					& 0.777		& \cite{socher2011semi} \\
	\hline
	MV-RNN				& 0.790		& \cite{socher2012semantic} \\
	\hline
	CCAE				& 0.778		& \cite{hermann2013role}
	\\ \hline
	NVSVM				& 0.794		& \multirow{2}{*}{\cite{wang2012baselines}} \\
	MNB					& 0.790		&   \\ 
	\hline
	G-dropout			& 0.790		& \multirow{2}{*}{\cite{wang2013fast}} \\
	F-dropout			& 0.791		&  \\
	\hline
	Tree-CRF			& 0.773		& \cite{nakagawa2010dependency} \\ 
	\hline
	Sent-Parser			& 0.795		& \cite{dong2015statistical} \\
	\hline
	\textbf{HWW2V with SVM (RBF)}			& \textbf{0.796} & \\
	\hline
\end{tabular}
	\label{review}
\end{table}

Some details on the methods presented in Table~\ref{review} are provided
below. 
In~\cite{kim2014convolutional} the authors report a series of experimental results using 
Convolutional Neural Networks (CNN) trained on top of pre-trained word vectors for 
sentence-level classification tasks. In~\cite{socher2011semi} a novel machine learning framework based 
on Recursive Autoencoders (RAE) for sentence-level prediction of sentiment label distributions, 
is introduced. The method supports learning vector space representations for multi-word phrases, 
without using any pre-defined sentiment lexica or polarity shifting rules. Also, in~\cite{socher2012semantic} 
a Recursive Neural Network model (RNN) is introduced that learns compositional vector 
representations for phrases and sentences of arbitrary syntactic type and length. 
Herman and Blunsom~\citeyear{hermann2013role} draw upon advances in the learning of vector space representations 
of sentential semantics and the transparent interface between syntax and semantics provided by 
Combinatory Categorial Grammar (CCG), and they introduce Combinatory Categorial Autoencoders (CCAE). 
This model leverages the CCG combinatory operators to guide a nonlinear transformation of 
meaning within a sentence. They use this model to learn high dimensional embeddings for 
sentences and evaluate them in a range of tasks, demonstrating that the incorporation of syntax 
allows a concise model to learn representations that are both effective and general.
Wand and Manning~\citeyear{wang2012baselines} proposed simple but novel 
Na{\"i}ve Bayes and SVM variants as feature values (MNB and NVSVM, respectively), which 
performed well across tasks and datasets, sometimes providing new state-of-the-art 
performance levels. In~\cite{wang2013fast} 
the dropout training model~\cite{hinton2012improving} was further improved
by randomly dropping out (zeroing) hidden units and input features during training of neural networks. 
This process speeds up training without losing classification performance.
Nakagawa et al.~\citeyear{nakagawa2010dependency} introduced a state-of-the-art 
dependency tree-based classification method that uses Conditional Random Fields (CRF) with hidden variables.
Dong et al.~\citeyear{dong2015statistical} presented a statistical parsing framework for sentence-level 
sentiment classification. Unlike previous works that employ syntactic parsing results for sentiment analysis, they develop a 
statistical parser to directly analyze the sentiment structure of a sentence.

Several classifiers have 
been tested in our HWW2V approach, where the SVM model with RBF kernel 
gave the best results in terms of accuracy, i.e. 79.6\%, improving further the accuracy obtained by Weighted Word2Vec 
(for vectors with 300 dimensions) and BoW represention. 
Based on our results, HWW2V methodology outperforms most of the compared approaches (Table~\label{review}), 
and it is of low cost in terms of computational time compared to deep architectures, something of a vital interest 
for practical applications. In future work we plan to apply the HWW2V methodology in more datasets and 
explore its accuracy and efficiency to different languages, e.g. Greek, in order to further examine its value.


\bigskip

\noindent\textbf{Acknowledgements}\\
This work has been supported by mSensis S.A. and the 
General Secretariat for Research and Technology (GSRT),
Programme for the Development of Industrial Research and Technology-PAVET, 
\emph{Deep Learning Methods in Sentiment Analysis} (Ref.~No.~1493-BET-2013). 
The ownership of all possible future IPRs, related in any way with this work,
belong solely to mSensis S.A. as it is stated to the respective contract between mSensis S.A. and GSRT. 
Authors do not claim ownership to any possible future IPRs that might be related in any way with this work.

\vskip 0.2in
\bibliography{mybib_updated}

\begin{thebibliography}{45}%
\makeatletter
\providecommand \@ifxundefined [1]{%
 \@ifx{#1\undefined}
}%
\providecommand \@ifnum [1]{%
 \ifnum #1\expandafter \@firstoftwo
 \else \expandafter \@secondoftwo
 \fi
}%
\providecommand \@ifx [1]{%
 \ifx #1\expandafter \@firstoftwo
 \else \expandafter \@secondoftwo
 \fi
}%
\providecommand \natexlab [1]{#1}%
\providecommand \enquote  [1]{``#1''}%
\providecommand \bibnamefont  [1]{#1}%
\providecommand \bibfnamefont [1]{#1}%
\providecommand \citenamefont [1]{#1}%
\providecommand \href@noop [0]{\@secondoftwo}%
\providecommand \href [0]{\begingroup \@sanitize@url \@href}%
\providecommand \@href[1]{\@@startlink{#1}\@@href}%
\providecommand \@@href[1]{\endgroup#1\@@endlink}%
\providecommand \@sanitize@url [0]{\catcode `\\12\catcode `\$12\catcode
  `\&12\catcode `\#12\catcode `\^12\catcode `\_12\catcode `\%12\relax}%
\providecommand \@@startlink[1]{}%
\providecommand \@@endlink[0]{}%
\providecommand \url  [0]{\begingroup\@sanitize@url \@url }%
\providecommand \@url [1]{\endgroup\@href {#1}{\urlprefix }}%
\providecommand \urlprefix  [0]{URL }%
\providecommand \Eprint [0]{\href }%
\providecommand \doibase [0]{http://dx.doi.org/}%
\providecommand \selectlanguage [0]{\@gobble}%
\providecommand \bibinfo  [0]{\@secondoftwo}%
\providecommand \bibfield  [0]{\@secondoftwo}%
\providecommand \translation [1]{[#1]}%
\providecommand \BibitemOpen [0]{}%
\providecommand \bibitemStop [0]{}%
\providecommand \bibitemNoStop [0]{.\EOS\space}%
\providecommand \EOS [0]{\spacefactor3000\relax}%
\providecommand \BibitemShut  [1]{\csname bibitem#1\endcsname}%
\let\auto@bib@innerbib\@empty
\bibitem [{\citenamefont {Pang}\ \emph {et~al.}(2002)\citenamefont {Pang},
  \citenamefont {Lee},\ and\ \citenamefont {Vaithyanathan}}]{pang2002thumbs}%
  \BibitemOpen
  \bibfield  {author} {\bibinfo {author} {\bibfnamefont {B.}~\bibnamefont
  {Pang}}, \bibinfo {author} {\bibfnamefont {L.}~\bibnamefont {Lee}}, \ and\
  \bibinfo {author} {\bibfnamefont {S.}~\bibnamefont {Vaithyanathan}},\ }in\
  \href@noop {} {\emph {\bibinfo {booktitle} {Proceedings of the ACL-02
  Conference on Empirical Methods in Natural Language Processing - Volume
  10}}},\ \bibinfo {series and number} {EMNLP '02}\ (\bibinfo  {publisher}
  {Association for Computational Linguistics},\ \bibinfo {year} {2002})\ pp.\
  \bibinfo {pages} {79--86}\BibitemShut {NoStop}%
\bibitem [{\citenamefont {Liu}(2012)}]{liu2012sentiment}%
  \BibitemOpen
  \bibfield  {author} {\bibinfo {author} {\bibfnamefont {B.}~\bibnamefont
  {Liu}},\ }\href@noop {} {\bibfield  {journal} {\bibinfo  {journal} {Synthesis
  Lectures on Human Language Technologies}\ }\textbf {\bibinfo {volume} {5}},\
  \bibinfo {pages} {1} (\bibinfo {year} {2012})}\BibitemShut {NoStop}%
\bibitem [{\citenamefont {Xu}\ \emph {et~al.}(2012)\citenamefont {Xu},
  \citenamefont {Peng},\ and\ \citenamefont {Cheng}}]{xu2012identifying}%
  \BibitemOpen
  \bibfield  {author} {\bibinfo {author} {\bibfnamefont {T.}~\bibnamefont
  {Xu}}, \bibinfo {author} {\bibfnamefont {Q.}~\bibnamefont {Peng}}, \ and\
  \bibinfo {author} {\bibfnamefont {Y.}~\bibnamefont {Cheng}},\ }\href@noop {}
  {\bibfield  {journal} {\bibinfo  {journal} {Knowledge-Based Systems}\
  }\textbf {\bibinfo {volume} {35}},\ \bibinfo {pages} {279} (\bibinfo {year}
  {2012})}\BibitemShut {NoStop}%
\bibitem [{\citenamefont {Hagenau}\ \emph {et~al.}(2013)\citenamefont
  {Hagenau}, \citenamefont {Liebmann},\ and\ \citenamefont
  {Neumann}}]{hagenau2013automated}%
  \BibitemOpen
  \bibfield  {author} {\bibinfo {author} {\bibfnamefont {M.}~\bibnamefont
  {Hagenau}}, \bibinfo {author} {\bibfnamefont {M.}~\bibnamefont {Liebmann}}, \
  and\ \bibinfo {author} {\bibfnamefont {D.}~\bibnamefont {Neumann}},\
  }\href@noop {} {\bibfield  {journal} {\bibinfo  {journal} {Decision Support
  Systems}\ }\textbf {\bibinfo {volume} {55}},\ \bibinfo {pages} {685}
  (\bibinfo {year} {2013})}\BibitemShut {NoStop}%
\bibitem [{\citenamefont {Maks}\ and\ \citenamefont
  {Vossen}(2012)}]{maks2012lexicon}%
  \BibitemOpen
  \bibfield  {author} {\bibinfo {author} {\bibfnamefont {I.}~\bibnamefont
  {Maks}}\ and\ \bibinfo {author} {\bibfnamefont {P.}~\bibnamefont {Vossen}},\
  }\href@noop {} {\bibfield  {journal} {\bibinfo  {journal} {Decision Support
  Systems}\ }\textbf {\bibinfo {volume} {53}},\ \bibinfo {pages} {680}
  (\bibinfo {year} {2012})}\BibitemShut {NoStop}%
\bibitem [{\citenamefont {Turney}(2002)}]{turney2002thumbs}%
  \BibitemOpen
  \bibfield  {author} {\bibinfo {author} {\bibfnamefont {P.~D.}\ \bibnamefont
  {Turney}},\ }in\ \href@noop {} {\emph {\bibinfo {booktitle} {Proceedings of
  the 40th Annual Meeting on Association for Computational Linguistics}}},\
  \bibinfo {series and number} {ACL '02}\ (\bibinfo  {publisher} {Association
  for Computational Linguistics},\ \bibinfo {year} {2002})\ pp.\ \bibinfo
  {pages} {417--424}\BibitemShut {NoStop}%
\bibitem [{\citenamefont {Hatzivassiloglou}\ and\ \citenamefont
  {McKeown}(1997)}]{hatzivassiloglou1997predicting}%
  \BibitemOpen
  \bibfield  {author} {\bibinfo {author} {\bibfnamefont {V.}~\bibnamefont
  {Hatzivassiloglou}}\ and\ \bibinfo {author} {\bibfnamefont {K.~R.}\
  \bibnamefont {McKeown}},\ }in\ \href@noop {} {\emph {\bibinfo {booktitle}
  {Proceedings of the 35th annual meeting of the association for computational
  linguistics and eighth conference of the european chapter of the association
  for computational linguistics}}}\ (\bibinfo {organization} {Association for
  Computational Linguistics},\ \bibinfo {year} {1997})\ pp.\ \bibinfo {pages}
  {174--181}\BibitemShut {NoStop}%
\bibitem [{\citenamefont {Wiebe}(2000)}]{wiebe2000learning}%
  \BibitemOpen
  \bibfield  {author} {\bibinfo {author} {\bibfnamefont {J.}~\bibnamefont
  {Wiebe}},\ }in\ \href@noop {} {\emph {\bibinfo {booktitle} {Proceedings of
  the Seventeenth National Conference on Artificial Intelligence and Twelfth
  Conference on Innovative Applications of Artificial Intelligence}}}\
  (\bibinfo  {publisher} {AAAI Press},\ \bibinfo {year} {2000})\ pp.\ \bibinfo
  {pages} {735--740}\BibitemShut {NoStop}%
\bibitem [{\citenamefont {Hu}\ and\ \citenamefont {Liu}(2004)}]{hu2004mining}%
  \BibitemOpen
  \bibfield  {author} {\bibinfo {author} {\bibfnamefont {M.}~\bibnamefont
  {Hu}}\ and\ \bibinfo {author} {\bibfnamefont {B.}~\bibnamefont {Liu}},\ }in\
  \href@noop {} {\emph {\bibinfo {booktitle} {Proceedings of the tenth ACM
  SIGKDD international conference on Knowledge discovery and data mining}}},\
  \bibinfo {series and number} {KDD '04}\ (\bibinfo {organization} {ACM},\
  \bibinfo {year} {2004})\ pp.\ \bibinfo {pages} {168--177}\BibitemShut
  {NoStop}%
\bibitem [{\citenamefont {Taboada}\ \emph {et~al.}(2006)\citenamefont
  {Taboada}, \citenamefont {Anthony},\ and\ \citenamefont
  {Voll}}]{taboada2006methods}%
  \BibitemOpen
  \bibfield  {author} {\bibinfo {author} {\bibfnamefont {M.}~\bibnamefont
  {Taboada}}, \bibinfo {author} {\bibfnamefont {C.}~\bibnamefont {Anthony}}, \
  and\ \bibinfo {author} {\bibfnamefont {K.}~\bibnamefont {Voll}},\ }in\
  \href@noop {} {\emph {\bibinfo {booktitle} {Proceedings of the Fifth
  International Conference on Language Resources and Evaluation}}},\ \bibinfo
  {series and number} {LREC '06}\ (\bibinfo {year} {2006})\ pp.\ \bibinfo
  {pages} {427--432}\BibitemShut {NoStop}%
\bibitem [{\citenamefont {Benamara}\ \emph {et~al.}(2007)\citenamefont
  {Benamara}, \citenamefont {Cesarano}, \citenamefont {Picariello},
  \citenamefont {Recupero},\ and\ \citenamefont
  {Subrahmanian}}]{benamara2007sentiment}%
  \BibitemOpen
  \bibfield  {author} {\bibinfo {author} {\bibfnamefont {F.}~\bibnamefont
  {Benamara}}, \bibinfo {author} {\bibfnamefont {C.}~\bibnamefont {Cesarano}},
  \bibinfo {author} {\bibfnamefont {A.}~\bibnamefont {Picariello}}, \bibinfo
  {author} {\bibfnamefont {D.~R.}\ \bibnamefont {Recupero}}, \ and\ \bibinfo
  {author} {\bibfnamefont {V.~S.}\ \bibnamefont {Subrahmanian}},\ }in\
  \href@noop {} {\emph {\bibinfo {booktitle} {Proceedings of International
  Conference on Weblogs and Social Media}}},\ \bibinfo {series and number}
  {ICWSM '10}\ (\bibinfo {year} {2007})\BibitemShut {NoStop}%
\bibitem [{\citenamefont {Taboada}\ \emph {et~al.}(2011)\citenamefont
  {Taboada}, \citenamefont {Brooke}, \citenamefont {Tofiloski}, \citenamefont
  {Voll},\ and\ \citenamefont {Stede}}]{taboada2011lexicon}%
  \BibitemOpen
  \bibfield  {author} {\bibinfo {author} {\bibfnamefont {M.}~\bibnamefont
  {Taboada}}, \bibinfo {author} {\bibfnamefont {J.}~\bibnamefont {Brooke}},
  \bibinfo {author} {\bibfnamefont {M.}~\bibnamefont {Tofiloski}}, \bibinfo
  {author} {\bibfnamefont {K.}~\bibnamefont {Voll}}, \ and\ \bibinfo {author}
  {\bibfnamefont {M.}~\bibnamefont {Stede}},\ }\href@noop {} {\bibfield
  {journal} {\bibinfo  {journal} {Computational linguistics}\ }\textbf
  {\bibinfo {volume} {37}},\ \bibinfo {pages} {267} (\bibinfo {year}
  {2011})}\BibitemShut {NoStop}%
\bibitem [{\citenamefont {Tong}(2001)}]{tong2001operational}%
  \BibitemOpen
  \bibfield  {author} {\bibinfo {author} {\bibfnamefont {R.~M.}\ \bibnamefont
  {Tong}},\ }in\ \href@noop {} {\emph {\bibinfo {booktitle} {Working Notes of
  the ACM SIGIR 2001 Workshop on Operational Text Classification}}},\
  Vol.~\bibinfo {volume} {1}\ (\bibinfo {year} {2001})\ p.~\bibinfo {pages}
  {6}\BibitemShut {NoStop}%
\bibitem [{\citenamefont {Turney}\ and\ \citenamefont
  {Littman}(2003)}]{turney2003measuring}%
  \BibitemOpen
  \bibfield  {author} {\bibinfo {author} {\bibfnamefont {P.~D.}\ \bibnamefont
  {Turney}}\ and\ \bibinfo {author} {\bibfnamefont {M.~L.}\ \bibnamefont
  {Littman}},\ }\href@noop {} {\bibfield  {journal} {\bibinfo  {journal} {ACM
  Transactions on Information Systems}\ }\textbf {\bibinfo {volume} {21}},\
  \bibinfo {pages} {315} (\bibinfo {year} {2003})}\BibitemShut {NoStop}%
\bibitem [{\citenamefont {Whitelaw}\ \emph {et~al.}(2005)\citenamefont
  {Whitelaw}, \citenamefont {Garg},\ and\ \citenamefont
  {Argamon}}]{whitelaw2005using}%
  \BibitemOpen
  \bibfield  {author} {\bibinfo {author} {\bibfnamefont {C.}~\bibnamefont
  {Whitelaw}}, \bibinfo {author} {\bibfnamefont {N.}~\bibnamefont {Garg}}, \
  and\ \bibinfo {author} {\bibfnamefont {S.}~\bibnamefont {Argamon}},\ }in\
  \href@noop {} {\emph {\bibinfo {booktitle} {Proceedings of the 14th ACM
  international conference on Information and knowledge management}}},\
  \bibinfo {series and number} {CIKM '05}\ (\bibinfo {organization} {ACM},\
  \bibinfo {year} {2005})\ pp.\ \bibinfo {pages} {625--631}\BibitemShut
  {NoStop}%
\bibitem [{\citenamefont {Martin}\ and\ \citenamefont
  {White}(2005)}]{martinlanguage}%
  \BibitemOpen
  \bibfield  {author} {\bibinfo {author} {\bibfnamefont {J.~R.}\ \bibnamefont
  {Martin}}\ and\ \bibinfo {author} {\bibfnamefont {P.~R.}\ \bibnamefont
  {White}},\ }\href@noop {} {\emph {\bibinfo {title} {The Language of
  Evaluation: Appraisal in English}}}\ (\bibinfo  {publisher} {Palgrave
  Macmillan},\ \bibinfo {address} {New York},\ \bibinfo {year}
  {2005})\BibitemShut {NoStop}%
\bibitem [{\citenamefont {Ye}\ \emph {et~al.}(2009)\citenamefont {Ye},
  \citenamefont {Zhang},\ and\ \citenamefont {Law}}]{ye2009sentiment}%
  \BibitemOpen
  \bibfield  {author} {\bibinfo {author} {\bibfnamefont {Q.}~\bibnamefont
  {Ye}}, \bibinfo {author} {\bibfnamefont {Z.}~\bibnamefont {Zhang}}, \ and\
  \bibinfo {author} {\bibfnamefont {R.}~\bibnamefont {Law}},\ }\href@noop {}
  {\bibfield  {journal} {\bibinfo  {journal} {Expert Systems with
  Applications}\ }\textbf {\bibinfo {volume} {36}},\ \bibinfo {pages} {6527}
  (\bibinfo {year} {2009})}\BibitemShut {NoStop}%
\bibitem [{\citenamefont {Prabowo}\ and\ \citenamefont
  {Thelwall}(2009)}]{prabowo2009sentiment}%
  \BibitemOpen
  \bibfield  {author} {\bibinfo {author} {\bibfnamefont {R.}~\bibnamefont
  {Prabowo}}\ and\ \bibinfo {author} {\bibfnamefont {M.}~\bibnamefont
  {Thelwall}},\ }\href@noop {} {\bibfield  {journal} {\bibinfo  {journal}
  {Journal of Informetrics}\ }\textbf {\bibinfo {volume} {3}},\ \bibinfo
  {pages} {143} (\bibinfo {year} {2009})}\BibitemShut {NoStop}%
\bibitem [{\citenamefont {Salton}\ and\ \citenamefont
  {McGill}(1986)}]{salton1986introduction}%
  \BibitemOpen
  \bibfield  {author} {\bibinfo {author} {\bibfnamefont {G.}~\bibnamefont
  {Salton}}\ and\ \bibinfo {author} {\bibfnamefont {M.~J.}\ \bibnamefont
  {McGill}},\ }\href@noop {} {\emph {\bibinfo {title} {Introduction to modern
  information retrieval}}}\ (\bibinfo  {publisher} {McGraw-Hill, Inc.},\
  \bibinfo {year} {1986})\BibitemShut {NoStop}%
\bibitem [{\citenamefont {Mikolov}\ \emph
  {et~al.}(2013{\natexlab{a}})\citenamefont {Mikolov}, \citenamefont {Chen},
  \citenamefont {Corrado},\ and\ \citenamefont {Dean}}]{mikolov2013efficient}%
  \BibitemOpen
  \bibfield  {author} {\bibinfo {author} {\bibfnamefont {T.}~\bibnamefont
  {Mikolov}}, \bibinfo {author} {\bibfnamefont {K.}~\bibnamefont {Chen}},
  \bibinfo {author} {\bibfnamefont {G.}~\bibnamefont {Corrado}}, \ and\
  \bibinfo {author} {\bibfnamefont {J.}~\bibnamefont {Dean}},\ }\href@noop {}
  {\bibfield  {journal} {\bibinfo  {journal} {arXiv preprint arXiv:1301.3781}\
  } (\bibinfo {year} {2013}{\natexlab{a}})}\BibitemShut {NoStop}%
\bibitem [{\citenamefont {Wiegand}\ \emph {et~al.}(2010)\citenamefont
  {Wiegand}, \citenamefont {Balahur}, \citenamefont {Roth}, \citenamefont
  {Klakow},\ and\ \citenamefont {Montoyo}}]{wiegand2010survey}%
  \BibitemOpen
  \bibfield  {author} {\bibinfo {author} {\bibfnamefont {M.}~\bibnamefont
  {Wiegand}}, \bibinfo {author} {\bibfnamefont {A.}~\bibnamefont {Balahur}},
  \bibinfo {author} {\bibfnamefont {B.}~\bibnamefont {Roth}}, \bibinfo {author}
  {\bibfnamefont {D.}~\bibnamefont {Klakow}}, \ and\ \bibinfo {author}
  {\bibfnamefont {A.}~\bibnamefont {Montoyo}},\ }in\ \href@noop {} {\emph
  {\bibinfo {booktitle} {Proceedings of the workshop on negation and
  speculation in natural language processing}}}\ (\bibinfo {organization}
  {Association for Computational Linguistics},\ \bibinfo {year} {2010})\ pp.\
  \bibinfo {pages} {60--68}\BibitemShut {NoStop}%
\bibitem [{\citenamefont {Mikolov}\ \emph
  {et~al.}(2013{\natexlab{b}})\citenamefont {Mikolov}, \citenamefont {Yih},\
  and\ \citenamefont {Zweig}}]{mikolov2013linguistic}%
  \BibitemOpen
  \bibfield  {author} {\bibinfo {author} {\bibfnamefont {T.}~\bibnamefont
  {Mikolov}}, \bibinfo {author} {\bibfnamefont {W.-t.}\ \bibnamefont {Yih}}, \
  and\ \bibinfo {author} {\bibfnamefont {G.}~\bibnamefont {Zweig}},\ }in\
  \href@noop {} {\emph {\bibinfo {booktitle} {Proceedings of the 2013
  Conference of the North American Chapter of the Association for Computational
  Linguistics: Human Language Technologies}}}\ (\bibinfo  {publisher}
  {Association for Computational Linguistics},\ \bibinfo {year} {2013})\ pp.\
  \bibinfo {pages} {746--751}\BibitemShut {NoStop}%
\bibitem [{\citenamefont {Zanzotto}\ \emph {et~al.}(2010)\citenamefont
  {Zanzotto}, \citenamefont {Korkontzelos}, \citenamefont {Fallucchi},\ and\
  \citenamefont {Manandhar}}]{zanzotto2010estimating}%
  \BibitemOpen
  \bibfield  {author} {\bibinfo {author} {\bibfnamefont {F.~M.}\ \bibnamefont
  {Zanzotto}}, \bibinfo {author} {\bibfnamefont {I.}~\bibnamefont
  {Korkontzelos}}, \bibinfo {author} {\bibfnamefont {F.}~\bibnamefont
  {Fallucchi}}, \ and\ \bibinfo {author} {\bibfnamefont {S.}~\bibnamefont
  {Manandhar}},\ }in\ \href@noop {} {\emph {\bibinfo {booktitle} {Proceedings
  of the 23rd International Conference on Computational Linguistics}}},\
  \bibinfo {series and number} {COLING '10}\ (\bibinfo {organization}
  {Association for Computational Linguistics},\ \bibinfo {year} {2010})\ pp.\
  \bibinfo {pages} {1263--1271}\BibitemShut {NoStop}%
\bibitem [{\citenamefont {Mitchell}\ and\ \citenamefont
  {Lapata}(2010)}]{mitchell2010composition}%
  \BibitemOpen
  \bibfield  {author} {\bibinfo {author} {\bibfnamefont {J.}~\bibnamefont
  {Mitchell}}\ and\ \bibinfo {author} {\bibfnamefont {M.}~\bibnamefont
  {Lapata}},\ }\href@noop {} {\bibfield  {journal} {\bibinfo  {journal}
  {Cognitive science}\ }\textbf {\bibinfo {volume} {34}},\ \bibinfo {pages}
  {1388} (\bibinfo {year} {2010})}\BibitemShut {NoStop}%
\bibitem [{\citenamefont {Grefenstette}\ \emph {et~al.}(2013)\citenamefont
  {Grefenstette}, \citenamefont {Dinu}, \citenamefont {Zhang}, \citenamefont
  {Sadrzadeh},\ and\ \citenamefont {Baroni}}]{grefenstette2013multi}%
  \BibitemOpen
  \bibfield  {author} {\bibinfo {author} {\bibfnamefont {E.}~\bibnamefont
  {Grefenstette}}, \bibinfo {author} {\bibfnamefont {G.}~\bibnamefont {Dinu}},
  \bibinfo {author} {\bibfnamefont {Y.-Z.}\ \bibnamefont {Zhang}}, \bibinfo
  {author} {\bibfnamefont {M.}~\bibnamefont {Sadrzadeh}}, \ and\ \bibinfo
  {author} {\bibfnamefont {M.}~\bibnamefont {Baroni}},\ }\href@noop {}
  {\bibfield  {journal} {\bibinfo  {journal} {arXiv preprint arXiv:1301.6939}\
  } (\bibinfo {year} {2013})}\BibitemShut {NoStop}%
\bibitem [{\citenamefont {Yessenalina}\ and\ \citenamefont
  {Cardie}(2011)}]{yessenalina2011compositional}%
  \BibitemOpen
  \bibfield  {author} {\bibinfo {author} {\bibfnamefont {A.}~\bibnamefont
  {Yessenalina}}\ and\ \bibinfo {author} {\bibfnamefont {C.}~\bibnamefont
  {Cardie}},\ }in\ \href@noop {} {\emph {\bibinfo {booktitle} {Proceedings of
  the Conference on Empirical Methods in Natural Language Processing}}},\
  \bibinfo {series and number} {EMNLP '11}\ (\bibinfo  {publisher} {Association
  for Computational Linguistics},\ \bibinfo {address} {Stroudsburg, PA, USA},\
  \bibinfo {year} {2011})\ pp.\ \bibinfo {pages} {172--182}\BibitemShut
  {NoStop}%
\bibitem [{\citenamefont {Socher}\ \emph {et~al.}(2013)\citenamefont {Socher},
  \citenamefont {Perelygin}, \citenamefont {Wu}, \citenamefont {Chuang},
  \citenamefont {Manning}, \citenamefont {Ng},\ and\ \citenamefont
  {Potts}}]{socher2013recursive}%
  \BibitemOpen
  \bibfield  {author} {\bibinfo {author} {\bibfnamefont {R.}~\bibnamefont
  {Socher}}, \bibinfo {author} {\bibfnamefont {A.}~\bibnamefont {Perelygin}},
  \bibinfo {author} {\bibfnamefont {J.~Y.}\ \bibnamefont {Wu}}, \bibinfo
  {author} {\bibfnamefont {J.}~\bibnamefont {Chuang}}, \bibinfo {author}
  {\bibfnamefont {C.~D.}\ \bibnamefont {Manning}}, \bibinfo {author}
  {\bibfnamefont {A.~Y.}\ \bibnamefont {Ng}}, \ and\ \bibinfo {author}
  {\bibfnamefont {C.}~\bibnamefont {Potts}},\ }in\ \href@noop {} {\emph
  {\bibinfo {booktitle} {Proceedings of the conference on empirical methods in
  natural language processing}}},\ \bibinfo {series} {EMNLP '13}, Vol.\
  \bibinfo {volume} {1631}\ (\bibinfo {year} {2013})\ p.\ \bibinfo {pages}
  {1642}\BibitemShut {NoStop}%
\bibitem [{\citenamefont {Le}\ and\ \citenamefont
  {Mikolov}(2014)}]{le2014distributed}%
  \BibitemOpen
  \bibfield  {author} {\bibinfo {author} {\bibfnamefont {Q.~V.}\ \bibnamefont
  {Le}}\ and\ \bibinfo {author} {\bibfnamefont {T.}~\bibnamefont {Mikolov}},\
  }in\ \href@noop {} {\emph {\bibinfo {booktitle} {Proceedings of the 31st
  International Conference on Machine Learning}}},\ \bibinfo {series and
  number} {ICML '14}\ (\bibinfo {year} {2014})\ pp.\ \bibinfo {pages}
  {1188--1196}\BibitemShut {NoStop}%
\bibitem [{\citenamefont {Stone}\ \emph {et~al.}(1966)\citenamefont {Stone},
  \citenamefont {Dunphy},\ and\ \citenamefont {Smith}}]{stone1966general}%
  \BibitemOpen
  \bibfield  {author} {\bibinfo {author} {\bibfnamefont {P.~J.}\ \bibnamefont
  {Stone}}, \bibinfo {author} {\bibfnamefont {D.~C.}\ \bibnamefont {Dunphy}}, \
  and\ \bibinfo {author} {\bibfnamefont {M.~S.}\ \bibnamefont {Smith}},\
  }\href@noop {} {\emph {\bibinfo {title} {The General Inquirer: A Computer
  Approach to Content Analysis}}}\ (\bibinfo  {publisher} {MIT press},\
  \bibinfo {year} {1966})\BibitemShut {NoStop}%
\bibitem [{\citenamefont {Esuli}\ and\ \citenamefont
  {Sebastiani}(2006)}]{esuli2006sentiwordnet}%
  \BibitemOpen
  \bibfield  {author} {\bibinfo {author} {\bibfnamefont {A.}~\bibnamefont
  {Esuli}}\ and\ \bibinfo {author} {\bibfnamefont {F.}~\bibnamefont
  {Sebastiani}},\ }in\ \href@noop {} {\emph {\bibinfo {booktitle} {Proceedings
  of the Fifth International Conference on Language Resources and
  Evaluation}}},\ \bibinfo {series and number} {LREC '06}\ (\bibinfo
  {publisher} {European Language Resources Association},\ \bibinfo {year}
  {2006})\ pp.\ \bibinfo {pages} {417--422}\BibitemShut {NoStop}%
\bibitem [{\citenamefont {Nigam}\ \emph {et~al.}(1999)\citenamefont {Nigam},
  \citenamefont {Lafferty},\ and\ \citenamefont {McCallum}}]{nigam1999using}%
  \BibitemOpen
  \bibfield  {author} {\bibinfo {author} {\bibfnamefont {K.}~\bibnamefont
  {Nigam}}, \bibinfo {author} {\bibfnamefont {J.}~\bibnamefont {Lafferty}}, \
  and\ \bibinfo {author} {\bibfnamefont {A.}~\bibnamefont {McCallum}},\ }in\
  \href@noop {} {\emph {\bibinfo {booktitle} {IJCAI-99 workshop on machine
  learning for information filtering}}},\ Vol.~\bibinfo {volume} {1}\ (\bibinfo
  {year} {1999})\ pp.\ \bibinfo {pages} {61--67}\BibitemShut {NoStop}%
\bibitem [{\citenamefont {Pietra}\ \emph {et~al.}(1997)\citenamefont {Pietra},
  \citenamefont {Pietra},\ and\ \citenamefont {Lafferty}}]{pietra1997inducing}%
  \BibitemOpen
  \bibfield  {author} {\bibinfo {author} {\bibfnamefont {S.~D.}\ \bibnamefont
  {Pietra}}, \bibinfo {author} {\bibfnamefont {V.~D.}\ \bibnamefont {Pietra}},
  \ and\ \bibinfo {author} {\bibfnamefont {J.}~\bibnamefont {Lafferty}},\
  }\href@noop {} {\bibfield  {journal} {\bibinfo  {journal} {IEEE Transactions
  on Pattern Analysis and Machine Intelligence}\ }\textbf {\bibinfo {volume}
  {19}},\ \bibinfo {pages} {380} (\bibinfo {year} {1997})}\BibitemShut
  {NoStop}%
\bibitem [{\citenamefont {Cortes}\ and\ \citenamefont
  {Vapnik}(1995)}]{cortes1995support}%
  \BibitemOpen
  \bibfield  {author} {\bibinfo {author} {\bibfnamefont {C.}~\bibnamefont
  {Cortes}}\ and\ \bibinfo {author} {\bibfnamefont {V.}~\bibnamefont
  {Vapnik}},\ }\href@noop {} {\bibfield  {journal} {\bibinfo  {journal}
  {Machine learning}\ }\textbf {\bibinfo {volume} {20}},\ \bibinfo {pages}
  {273} (\bibinfo {year} {1995})}\BibitemShut {NoStop}%
\bibitem [{Note1()}]{Note1}%
  \BibitemOpen
  \bibinfo {note} {\protect \url
  {http://www.cs.cornell.edu/people/pabo/movie-review-data/rt-polaritydata.tar.gz}}\BibitemShut
  {NoStop}%
\bibitem [{\citenamefont {Pang}\ and\ \citenamefont {Lee}(2005)}]{pang2005a}%
  \BibitemOpen
  \bibfield  {author} {\bibinfo {author} {\bibfnamefont {B.}~\bibnamefont
  {Pang}}\ and\ \bibinfo {author} {\bibfnamefont {L.}~\bibnamefont {Lee}},\
  }in\ \href@noop {} {\emph {\bibinfo {booktitle} {Proceedings of the 43rd
  Annual Meeting on Association for Computational Linguistics}}},\ \bibinfo
  {series and number} {ACL '05}\ (\bibinfo  {publisher} {Association for
  Computational Linguistics},\ \bibinfo {year} {2005})\ pp.\ \bibinfo {pages}
  {115--124}\BibitemShut {NoStop}%
\bibitem [{Note2()}]{Note2}%
  \BibitemOpen
  \bibinfo {note} {\protect \url {http://scikit-learn.org}}\BibitemShut
  {NoStop}%
\bibitem [{\citenamefont {Kim}(2014)}]{kim2014convolutional}%
  \BibitemOpen
  \bibfield  {author} {\bibinfo {author} {\bibfnamefont {Y.}~\bibnamefont
  {Kim}},\ }in\ \href@noop {} {\emph {\bibinfo {booktitle} {Proceedings of the
  2014 Conference on Empirical Methods in Natural Language Processing
  (EMNLP)}}}\ (\bibinfo  {publisher} {Association for Computational
  Linguistics},\ \bibinfo {address} {Doha, Qatar},\ \bibinfo {year} {2014})\
  pp.\ \bibinfo {pages} {1746--1751}\BibitemShut {NoStop}%
\bibitem [{\citenamefont {Socher}\ \emph {et~al.}(2011)\citenamefont {Socher},
  \citenamefont {Pennington}, \citenamefont {Huang}, \citenamefont {Ng},\ and\
  \citenamefont {Manning}}]{socher2011semi}%
  \BibitemOpen
  \bibfield  {author} {\bibinfo {author} {\bibfnamefont {R.}~\bibnamefont
  {Socher}}, \bibinfo {author} {\bibfnamefont {J.}~\bibnamefont {Pennington}},
  \bibinfo {author} {\bibfnamefont {E.~H.}\ \bibnamefont {Huang}}, \bibinfo
  {author} {\bibfnamefont {A.~Y.}\ \bibnamefont {Ng}}, \ and\ \bibinfo {author}
  {\bibfnamefont {C.~D.}\ \bibnamefont {Manning}},\ }in\ \href@noop {} {\emph
  {\bibinfo {booktitle} {Proceedings of the Conference on Empirical Methods in
  Natural Language Processing}}},\ \bibinfo {series and number} {EMNLP '11}\
  (\bibinfo {organization} {Association for Computational Linguistics},\
  \bibinfo {year} {2011})\ pp.\ \bibinfo {pages} {151--161}\BibitemShut
  {NoStop}%
\bibitem [{\citenamefont {Socher}\ \emph {et~al.}(2012)\citenamefont {Socher},
  \citenamefont {Huval}, \citenamefont {Manning},\ and\ \citenamefont
  {Ng}}]{socher2012semantic}%
  \BibitemOpen
  \bibfield  {author} {\bibinfo {author} {\bibfnamefont {R.}~\bibnamefont
  {Socher}}, \bibinfo {author} {\bibfnamefont {B.}~\bibnamefont {Huval}},
  \bibinfo {author} {\bibfnamefont {C.~D.}\ \bibnamefont {Manning}}, \ and\
  \bibinfo {author} {\bibfnamefont {A.~Y.}\ \bibnamefont {Ng}},\ }in\
  \href@noop {} {\emph {\bibinfo {booktitle} {Proceedings of the 2012 Joint
  Conference on Empirical Methods in Natural Language Processing and
  Computational Natural Language Learning}}},\ \bibinfo {series and number}
  {EMNLP-CoNLL '12}\ (\bibinfo  {publisher} {Association for Computational
  Linguistics},\ \bibinfo {year} {2012})\ pp.\ \bibinfo {pages}
  {1201--1211}\BibitemShut {NoStop}%
\bibitem [{\citenamefont {Hermann}\ and\ \citenamefont
  {Blunsom}(2013)}]{hermann2013role}%
  \BibitemOpen
  \bibfield  {author} {\bibinfo {author} {\bibfnamefont {K.~M.}\ \bibnamefont
  {Hermann}}\ and\ \bibinfo {author} {\bibfnamefont {P.}~\bibnamefont
  {Blunsom}},\ }in\ \href@noop {} {\emph {\bibinfo {booktitle} {In Proceedings
  of the 51st Annual Meeting of the Association for Computational
  Linguistics}}}\ (\bibinfo {year} {2013})\ pp.\ \bibinfo {pages}
  {894--904}\BibitemShut {NoStop}%
\bibitem [{\citenamefont {Wang}\ and\ \citenamefont
  {Manning}(2012)}]{wang2012baselines}%
  \BibitemOpen
  \bibfield  {author} {\bibinfo {author} {\bibfnamefont {S.}~\bibnamefont
  {Wang}}\ and\ \bibinfo {author} {\bibfnamefont {C.~D.}\ \bibnamefont
  {Manning}},\ }in\ \href@noop {} {\emph {\bibinfo {booktitle} {Proceedings of
  the 50th Annual Meeting of the Association for Computational Linguistics:
  Short Papers - Volume 2}}},\ \bibinfo {series and number} {ACL '12}\
  (\bibinfo  {publisher} {Association for Computational Linguistics},\ \bibinfo
  {year} {2012})\ pp.\ \bibinfo {pages} {90--94}\BibitemShut {NoStop}%
\bibitem [{\citenamefont {Wang}\ and\ \citenamefont
  {Manning}(2013)}]{wang2013fast}%
  \BibitemOpen
  \bibfield  {author} {\bibinfo {author} {\bibfnamefont {S.}~\bibnamefont
  {Wang}}\ and\ \bibinfo {author} {\bibfnamefont {C.}~\bibnamefont {Manning}},\
  }in\ \href@noop {} {\emph {\bibinfo {booktitle} {Proceedings of the 30th
  International Conference on Machine Learning}}},\ \bibinfo {series and
  number} {ICML '13}\ (\bibinfo {year} {2013})\ pp.\ \bibinfo {pages}
  {118--126}\BibitemShut {NoStop}%
\bibitem [{\citenamefont {Nakagawa}\ \emph {et~al.}(2010)\citenamefont
  {Nakagawa}, \citenamefont {Inui},\ and\ \citenamefont
  {Kurohashi}}]{nakagawa2010dependency}%
  \BibitemOpen
  \bibfield  {author} {\bibinfo {author} {\bibfnamefont {T.}~\bibnamefont
  {Nakagawa}}, \bibinfo {author} {\bibfnamefont {K.}~\bibnamefont {Inui}}, \
  and\ \bibinfo {author} {\bibfnamefont {S.}~\bibnamefont {Kurohashi}},\ }in\
  \href@noop {} {\emph {\bibinfo {booktitle} {Human Language Technologies: The
  2010 Annual Conference of the North American Chapter of the Association for
  Computational Linguistics}}}\ (\bibinfo {organization} {Association for
  Computational Linguistics},\ \bibinfo {year} {2010})\ pp.\ \bibinfo {pages}
  {786--794}\BibitemShut {NoStop}%
\bibitem [{\citenamefont {Dong}\ \emph {et~al.}(2015)\citenamefont {Dong},
  \citenamefont {Wei}, \citenamefont {Liu}, \citenamefont {Zhou},\ and\
  \citenamefont {Xu}}]{dong2015statistical}%
  \BibitemOpen
  \bibfield  {author} {\bibinfo {author} {\bibfnamefont {L.}~\bibnamefont
  {Dong}}, \bibinfo {author} {\bibfnamefont {F.}~\bibnamefont {Wei}}, \bibinfo
  {author} {\bibfnamefont {S.}~\bibnamefont {Liu}}, \bibinfo {author}
  {\bibfnamefont {M.}~\bibnamefont {Zhou}}, \ and\ \bibinfo {author}
  {\bibfnamefont {K.}~\bibnamefont {Xu}},\ }\href@noop {} {\bibfield  {journal}
  {\bibinfo  {journal} {Computational Linguistics}\ }\textbf {\bibinfo {volume}
  {41}},\ \bibinfo {pages} {293} (\bibinfo {year} {2015})}\BibitemShut
  {NoStop}%
\bibitem [{\citenamefont {Hinton}\ \emph {et~al.}(2012)\citenamefont {Hinton},
  \citenamefont {Srivastava}, \citenamefont {Krizhevsky}, \citenamefont
  {Sutskever},\ and\ \citenamefont {Salakhutdinov}}]{hinton2012improving}%
  \BibitemOpen
  \bibfield  {author} {\bibinfo {author} {\bibfnamefont {G.~E.}\ \bibnamefont
  {Hinton}}, \bibinfo {author} {\bibfnamefont {N.}~\bibnamefont {Srivastava}},
  \bibinfo {author} {\bibfnamefont {A.}~\bibnamefont {Krizhevsky}}, \bibinfo
  {author} {\bibfnamefont {I.}~\bibnamefont {Sutskever}}, \ and\ \bibinfo
  {author} {\bibfnamefont {R.~R.}\ \bibnamefont {Salakhutdinov}},\ }\href@noop
  {} {\bibfield  {journal} {\bibinfo  {journal} {arXiv preprint
  arXiv:1207.0580}\ } (\bibinfo {year} {2012})}\BibitemShut {NoStop}%
\end{thebibliography}%

\end{document}